# DAILOC: Domain-Incremental Learning for Indoor Localization using Smartphones

Akhil Singampalli, Danish Gufran, Sudeep Pasricha
Department of Electrical and Computer Engineering
Colorado State University, Fort Collins, CO, USA
{akhil.singampalli, danish.gufran, sudeep}@colostate.edu

*Abstract*—Wi-Fi fingerprinting-based indoor localization faces significant challenges in real-world deployments due to domain shifts arising from device heterogeneity and temporal variations within indoor environments. Existing approaches often address these issues independently, resulting in poor generalization and susceptibility to catastrophic forgetting over time. In this work, we propose *DAILOC*, a novel domain-incremental learning framework that jointly addresses both temporal and device-induced domain shifts. *DAILOC* introduces a novel disentanglement strategy that separates domain shifts from location-relevant features using a multi-level variational autoencoder. Additionally, we introduce a novel memory-guided class latent alignment mechanism to address the effects of catastrophic forgetting over time. Experiments across multiple smartphones, buildings, and time instances demonstrate that *DAILOC* significantly outperforms state-of-the-art methods, achieving up to 2.74× lower average error and 4.6× lower worst-case error.

## I. Introduction

The indoor localization market is projected to reach USD 84.17 billion by 2030, as demand rises for precise positioning [1]. While GPS/GNSS has transformed outdoor positioning, indoor localization remains a difficult problem due to the absence of satellite signals and the complexity of indoor environments. Indoor Localization aims to estimate the location of a device or user within a building, providing crucial functionality in homes, offices, shopping centers, hospitals, and industrial settings. A widely used approach for indoor localization is Wi-Fi-based Received Signal Strength (RSS) fingerprinting [2], [3]. In this approach, during an offline phase, WiFi RSS measurements are collected at fixed locations (aka Reference Points or RPs) to build a fingerprint database. Often, a machine learning (ML) model is trained to map these fingerprints to known locations. In the online phase, the model uses real-time RSS measurements to predict the user's position based on the previously learned mappings.

However, in real-world deployments indoor environments are rarely static. Factors such as moving furniture, changing layouts, or shifting occupancy patterns can significantly alter signal characteristics [4]. This mismatch between training and deployment conditions, known as *domain shift*, arises from two major factors in indoor localization: temporal variations (e.g., furniture rearrangement, occupancy changes, seasonal effects) and device heterogeneity (e.g., different smartphones with varying hardware and firmware). These shifts disrupt signal patterns, causing ML models trained offline to degrade significantly over time or across devices.

The impact of such domain shifts is illustrated in Fig. 1. Here, the state-of-the-art CNRP model [19] is trained on data collected from 60 RPs using a BLU Vivo 8 (BLU) device at time = 0 and evaluated on the BLU, Samsung Galaxy S7 (S7), and HTC U11 (HTC) devices over time. The left plot shows that even on the training device (BLU), localization error increases sharply over time, demonstrating the effect of temporal domain shift. The right plot shows the model's performance on the three different devices at the same time point as the training instance, highlighting a domain shift caused by device heterogeneity. These results show how domain shifts, whether due to time or hardware, can significantly degrade model reliability in real-world settings.

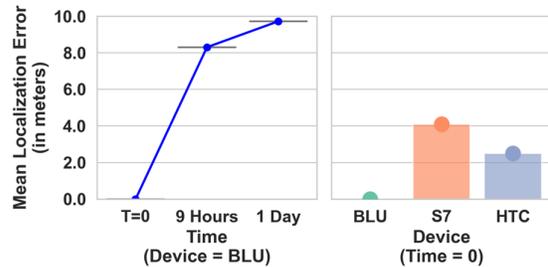

Fig. 1 Illustration on impact of domain shifts arising from device heterogeneity and temporal variations on indoor localization performance.

Current solutions typically rely on either retraining the model whenever accuracy drops, which requires costly re-collection of labeled data, or using adaptive learning strategies such as Federated Learning (FL) [5], where the model's own predictions are treated as labels for incoming unlabeled data. While these adaptive learning methods reduce the need for manual labeling, they introduce two major challenges: pseudo-label noise [6], where incorrect predictions are treated as ground truth during adaptation, and catastrophic forgetting [7], where learning on new data causes the model to lose previously acquired knowledge about earlier environments or known devices. These challenges make it difficult to sustain reliable indoor localization accuracy under persistent temporal and device-related domain shifts.

To address these important challenges posed by temporal and device domain shifts, we propose *DAILOC*, a novel domain-incremental learning framework for efficient WiFi-fingerprinting based indoor localization in dynamic indoor environments. The key contributions of this work are:

- A domain-incremental machine learning framework (*DAILOC*) that addresses dynamic device and temporal (environment) shifts by enabling continual adaptation.
- A disentangled representation learning strategy which improves generalization while reducing pseudo-label noise during unsupervised adaptation.

- A memory-guided class latent alignment mechanism that mitigates catastrophic forgetting by statistically aligning latent embeddings across sequential domains.
- Extensive evaluations on real-world RSS data, demonstrating improved accuracy and stability compared to state-of-the-art frameworks.

## II. RELATED WORK

Wi-Fi fingerprinting-based indoor localization has gained significant prominence in recent competitions hosted by IEEE IPIN [8] and Microsoft [9]. Early ML-based approaches, including KNN [10], HMM [11], and GPC [12], aimed to address fluctuations in RSS caused by human movement, and multipath effects but struggled to generalize across diverse environments due to their shallow architectures [13]. More recent deep learning (DL) models such as DNNLOC [14], CNNLOC [15], and ANVIL [16] have improved localization accuracy across diverse environments through learned feature extraction. Among these, ANVIL stands out by providing calibration-free and device-invariant localization using attention-based neural networks. However, ANVIL and other models are trained once and deployed statically, lacking the ability to adapt to unknown devices or environmental changes over time. This leaves them vulnerable to performance degradation in dynamic, real-world settings.

To address concerns with real-world deployment, federated learning (FL) approaches have been explored as a means to enable models to adapt continuously while preserving user privacy. For example, FedLoc [17] applies the popular FedAvg algorithm to collaboratively train a shared global model across heterogeneous clients, while FedHIL [18] introduces impact-driven aggregation to better handle client heterogeneity. CNRP [19] reduces reliance on labeled data through pseudo-labeling but suffer from noisy labels and unstable updates under domain shifts. In contrast, our approach in this work leverages pseudo-labels with strategies to suppress noise and enable long-term robust continual adaptation across evolving devices and indoor environments.

Beyond pseudo-labeling, a few domain adaptation techniques have been explored to handle domain shifts by aligning feature spaces between a source and target domain. Early approaches, such as signal strength difference and hyperbolic location fingerprinting [20], focused on calibration-free signal-level adjustments to reduce device heterogeneity. More recent deep learning-based methods aim to learn domain-invariant representations directly from data. For example, FIDORA [21] employs VAE-based augmentation and joint reconstruction-classification to improve adaptation with minimal labeling, while DAFI [22] aligns both marginal and conditional distributions for device-free localization. STELLAR [23] uses Siamese attention networks to handle temporal variations. TransLoc [24] introduces a cross-domain mapping framework that refines the source domain and constructs a shared latent space for heterogeneous feature transfer. While effective for isolated source–target settings, these methods are primarily designed for static, pairwise adaptation and do not scale naturally to real-world deployments where new devices, environments, and users emerge continuously.

While domain-incremental learning has been explored in vision and time-series applications [25], [26], it remains underexplored in indoor localization. Recent work such as DARE [27] highlights the importance of mitigating representation drift during domain-incremental learning to prevent catastrophic forgetting. While DARE focuses on generic task sequences, it assumes access to clean task boundaries and labeled data, conditions that rarely hold in real-world indoor localization. This limitation highlights the need for a framework like *DAILOC* that supports incremental domain adaptation, sustaining localization accuracy over time despite device heterogeneity and temporal drift.

## III. BACKGROUND: DOMAIN INCREMENTAL LEARNING FOR INDOOR LOCALIZATION

In real-world indoor localization systems, the relationship between Wi-Fi RSS and physical locations is sensitive to environmental changes. Over time, factors such as furniture movement, new access points (APs), or device variations alter the RSS patterns. This mismatch between training conditions and deployment conditions is known as domain shift. Unlike random noise, domain shifts represent deeper changes in signal distribution that can cause models to fail after deployment. Traditional domain adaptation methods try to adjust models trained in one setting (such as on a specific device or at a specific time) to work in another setting (like a different device or a later time). However, these approaches usually assume a one-time adaptation between two fixed domains. In practice, domain shifts are continuous and unpredictable within indoor localization, as unknown devices may appear at any time, and indoor environments may also evolve gradually or suddenly, in different ways.

To address these challenges, our *DAILOC* framework implements a domain-incremental learning strategy, enabling continual adaptation without retraining from scratch. During deployment, *DAILOC* follows two adaptation paths depending on the device's status. For an unknown device, labeled data is used for supervised onboarding. For a known device, unlabeled data from users is used for unsupervised adaptation, relying on pseudo-labels generated by the model itself. This setup reflects real-world deployment constraints, where collecting labeled samples for an unknown device once (e.g., by the localization solution provider) for onboarding is feasible. However, continuously collecting and manually labeling data across all time periods, environments, and devices quickly becomes impractical. Additionally, storing large amounts of data from all past domains would violate privacy guarantees and impose heavy memory and management overheads, especially in resource-limited mobile systems. Thus, the model must adapt incrementally using only current data, without retaining full access to prior domain samples. This realistic setup introduces two key challenges:

1. *Catastrophic forgetting:* where adapting to new unknown devices degrades prior knowledge.
2. *Pseudo-label noise:* where domain shifts cause incorrect self-predictions on known devices, harming model updates during unsupervised adaptation.

To address these challenges, it is essential to disentangle domain-specific variations from location-relevant features. Disentanglement isolates factors such as device biases and temporal environmental changes, allowing the model to focus adaptation on domain-specific components while preserving invariant features critical for reliable localization across evolving conditions. In the next section, we discuss *DAILOC*, with details of how it implements these principles using multi-level variational autoencoder (ML-VAE), disentangled representation, and memory-guided class latent alignment.

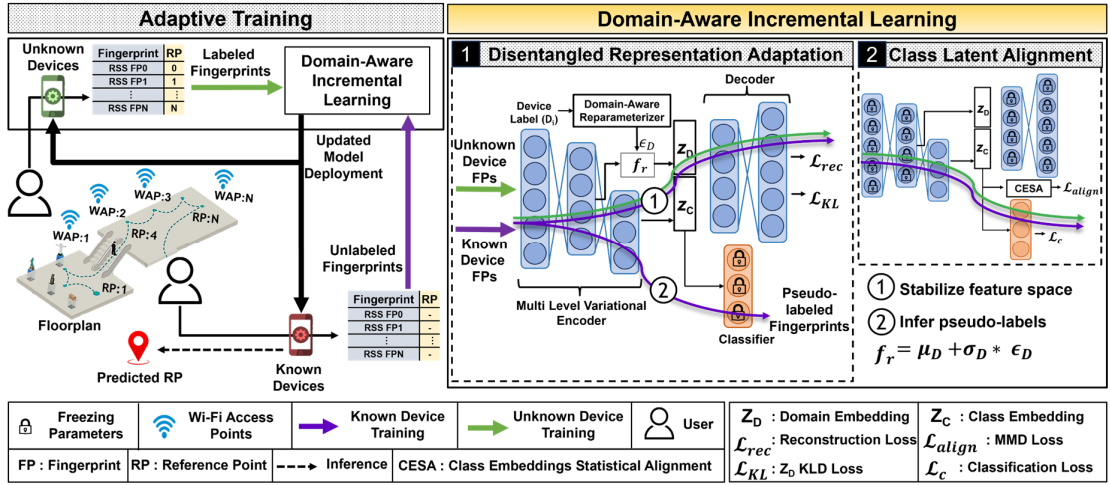

Fig. 2 Overview of the *DAILOC* framework using domain-aware incremental learning for robust Wi-Fi RSS fingerprinting-based indoor localization

## IV. *DAILOC* FRAMEWORK

To meet the challenges of operating under dynamic conditions, where unknown devices appear and indoor environments evolve over time, we propose *DAILOC*, a domain-incremental learning framework (Fig. 2) that enables models to adapt continually without requiring access to earlier domain data. *DAILOC* performs adaptive training which consists of two main stages: offline pretraining and online domain-aware incremental learning.

In the offline phase, *DAILOC* is trained on a curated labeled dataset, where RSS fingerprints are collected from various RPs using a fixed device under controlled conditions. These fingerprints are mapped to known physical coordinates (RPs) and used to learn location-discriminative features. At the core of DAILOC is a multi-level variational autoencoder (ML-VAE) comprising an encoder and decoder, and paired with a lightweight classifier. The encoder extracts latent representations from RSS, the decoder reconstructs the inputs to retain signal characteristics, and the classifier maps location-relevant features to RPs. The model at this stage is jointly optimized to capture general RSS-location relationships under relatively stable device and environmental condition providing a strong initialization before encountering the domain variability present during deployment. Once deployed in the online phase, *DAILOC* continuously collects crowdsourced RSS fingerprints from end-user devices and performs real-time localization. Unlike the curated offline dataset, these online samples often lack ground-truth location labels and may originate from unknown devices or environments undergoing gradual changes.

To adapt to dynamic domain data, *DAILOC* includes a domain-aware hybrid incremental learning pipeline composed of two key stages: 1) Disentangled Representation Learning and 2) Class Latent Alignment. In the first stage, incoming RSS fingerprints are encoded through the ML-VAE that separates the features into location-relevant and domain-specific factors. A domain-aware reparameterization mechanism ensures consistent modeling of domain variations while preserving localization-relevant information. Once the domain feature space is stabilized, pseudo-labels are generated for unlabeled data from users using a frozen classifier. By adapting the encoder before label generation, *DAILOC* reduces pseudo-label noise caused by domain shifts. In the second stage, a Class Embedding Statistical Alignment (CESA) module aligns the pseudo-labeled latent representations from the current domain with prototypes stored from previous domains. This alignment is guided by a Maximum Mean Discrepancy (MMD) loss to maintain consistency without accessing previous raw data.

Through this two-stage disentanglement and alignment based incremental learning pipeline, *DAILOC* achieves robust domain-aware incremental learning across temporal and device shifts without explicit retraining. Each stage of this incremental learning pipeline is described next.

### A. Disentangled Representation Adaptation

Effective domain-incremental adaptation requires separating device and environment-specific variations from stable location-relevant features. To achieve this, *DAILOC* employs a ML-VAE that disentangles input RSS fingerprints from both known and unknown devices into two latent spaces: $z_C$ for capturing domain-invariant, location-relevant class features and $z_D$ for capturing domain-specific variations arising from device or environmental shifts (Fig. 3). This disentangled representation allows adaptation to new domains by adjusting for signal variations while preserving location information critical for accurate localization.

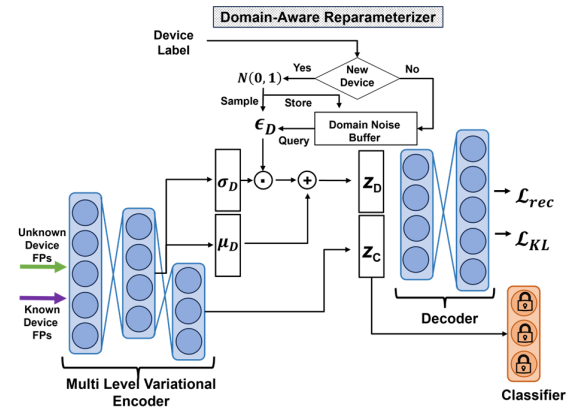

Fig. 3 *DAILOC*'s multi-level VAE with disentanglement representation

To achieve disentanglement as shown in Fig. 3, a domain-aware reparameterization is used inside the ML-VAE. Inspired by grouped observation frameworks such as [28], the encoder predicts mean vector $\mu_D$ and standard deviation $\sigma_D$ for the domain-specific latent space, but instead of applying random noise independently for each sample, a fixed noise vector $\epsilon_D$ for each domain $D$ is stored in a Domain Noise Buffer. During reparameterization, the corresponding $\epsilon_D$ is queried and applied to compute the domain-specific latent representation, as shown in (1):

$$z_D = \mu_D + \sigma_D * \epsilon_D \quad (1)$$

where $\epsilon_D \sim N(0,1)$. This consistent sampling forces $z_D$ to capture coherent domain-level variations such as device biases or temporal drifts. To regularize the latent spaces, we apply two loss components. First, a Kullback–Leibler (KL) divergence loss is used to encourage the learned posterior $N(\mu, \sigma^2)$ to remain close to a standard normal prior $N(0, I)$ promoting smoothness and compactness, as defined in (2):

$$\mathcal{L}_{KL} = D_{KL}(N(\mu, \sigma^2) \parallel N(0, I)) \quad (2)$$

Second, a reconstruction loss $\mathcal{L}_{rec}$ is used to minimize the difference between the input RSS fingerprint $x$ and its reconstruction $\hat{x}$ from the concatenated latent representations $[z_D, z_C]$ as defined in (3):

$$\mathcal{L}_{rec} = \|x - \hat{x}\|^2 \quad (3)$$

This two-part regularization leads to tight intra-domain clusters in $z_D$, while $z_C$ remains free from such structured noise and captures stable location-relevant features.

During adaptation, the classifier is frozen (see Fig. 3), and only the encoder is updated. This ensures that $z_D$ gradually adapts to evolving domains while preserving the existing decision boundaries in the classifier. Once the feature space stabilizes, pseudo-labels are inferred for unlabeled data from known devices using the frozen classifier to serve as inputs to the Class Latent Alignment stage (discussed next). By isolating domain-specific noise in $z_D$ and freezing the classifier, DAILOC aligns its feature space before any label-driven updates occur, significantly reducing the risk of pseudo-label noise. This controlled adaptation enables the model to refine its representations without corrupting learned location mappings, resulting in improved generalization to device and temporal domain shifts while supporting seamless unsupervised adaptation.

### B. Class Latent Alignment

To prevent catastrophic forgetting and preserve location identity across evolving domains, DAILOC introduces a Class Embedding Statistical Alignment (CESA) mechanism, as illustrated in Fig. 4. CESA aligns class embeddings from the current domain with historical prototypes which are location-relevant latent vectors $z_r$ stored during supervised onboarding of new devices in a Representation Memory Buffer. When an unknown device is introduced, supervised onboarding is performed to generate labeled location-relevant representations $z_C$. Instead of storing all samples, reservoir sampling is used, which is a random selection process that guarantees maintaining at least one representative embedding $z_C$ per reference point (RP). For known devices, similarity between the current pseudo-labeled embeddings $z_C$ and historical prototypes $z_r$ is enforced using a MMD loss as defined in (4):

$$\mathcal{L}_{align} = \left\| \frac{1}{n} \sum_{i=1}^{n} \phi(z_{c,i}) - \frac{1}{m} \sum_{i=1}^{m} \phi(z_{r,j}) \right\|_H^2 \quad (4)$$

where $n$ and $m$ represent the number of embeddings from the current domain and the reference memory buffer, respectively, and $\phi(.)$ is the mapping to a reproducing kernel Hilbert space $H$, enabling the comparison of distributions in a high-dimensional feature space. This loss minimizes the distributional discrepancy between current and historical embeddings preventing catastrophic forgetting. During this alignment phase, the encoder and decoder remains frozen, ensuring that latent representations do not drift. Only the classifier is updated using pseudo-labeled samples with a supervised classification loss ($\mathcal{L}_c$).

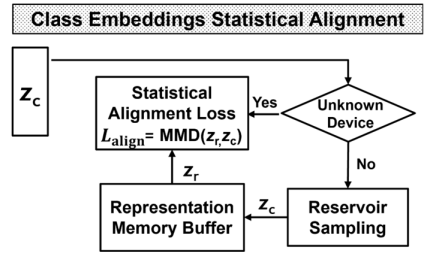

Fig. 4 Statistical alignment of class embeddings using MMD loss to prevent catastophic forgetting, as part of the CESA module.

### C. Joint Domain-Aware Optimization

The synergy between memory-based alignment and disentangled representation ensures that DAILOC maintains the plasticity needed to adapt to unknown devices and the stability required to retain prior knowledge of know devices, while also supporting continuous adaptation to temporal shifts. During adaptation, the outputs from both stages are optimized together. In the disentanglement stage, the reconstruction loss $\mathcal{L}_{rec}$ and KL divergence loss $\mathcal{L}_{KL}$ are minimized to stabilize and regularize the latent representations. After pseudo-labels are generated, the class alignment stage minimizes the MMD-based alignment loss $\mathcal{L}_{align}$ and the classification loss $\mathcal{L}_c$ to align embeddings and refine decision boundaries. These losses are summed up to form the total objective loss:

$$\mathcal{L}_{total} = \mathcal{L}_{rec} + \mathcal{L}_{KL} + \mathcal{L}_{align} + \mathcal{L}_c \quad (5)$$

This joint optimization enables DAILOC to mitigate catastrophic forgetting and achieve seamless, long-term integration of devices and environments under dynamic conditions. After adaptation, the updated encoder and classifier are deployed back for inference, ensuring improved localization performance in future predictions.

## V. EXPERIMENTAL RESULTS

### A. Experimental Setup

We evaluate DAILOC in a realistic continual learning scenario under diverse scenarios that reflects both device heterogeneity and temporal signal drifts. Experiments were conducted in two buildings with distinct physical layouts and material compositions. Building 1 consists of 60 RPs and up to 193 visible WiFi APs, featuring wood-cement construction, spacious open areas, and office infrastructure. Building 2 consists of 48 RPs and up to 168 WiFi APs, with a modern metal–wood layout with dense shelving and heavy equipment, leading to strong signal reflection and attenuation effects. RPs were placed at 1-meter intervals to ensure fine-grained localization. RSS values were standardized to the range [0, –100 dBm] for consistency across devices and locations.

The DAILOC architecture for experiments is based on ML-VAE model paired with a lightweight classifier, jointly optimized to disentangle location-relevant and domain-specific features. The encoder consists of two hidden layers with 256 and 128 neurons (with ReLU activations), followed by two separate branches: one that generates the mean and variance for the 16-dimensional latent representation $z_D$, and the other that maps to the 16-dimensional domain-invariant latent representation $z_C$ via a 64-unit intermediate layer. The decoder reconstructs input RSS vectors from the concatenated latent representations $[z_D, z_C]$ using two dense layers with 128 and 256 neurons, followed by a sigmoid output layer. The classifier operates on $z_C$ and includes one 64-unit hidden layer and a SoftMax output layer for RP classification. The encoder, decoder, and classifier contain approximately 88K, 80K, and 89K parameters respectively. Training uses the Adam optimizer with a learning rate of $1\times10^{-3}$. Representation memory buffer size $M$ is set to be equal to the number of RPs to accommodate at least one class prototype per RP.

To simulate device-level domain shifts, data was collected using six smartphones from different manufacturers (see Table 1). Devices varied in both hardware (Wi-Fi chipsets) and software (firmware-level signal processing). Even when chipsets were identical, manufacturer-specific differences in filtering introduced significant signal variation, creating diverse hardware and software heterogeneity conditions. To capture temporal dynamics, data was collected over six time instances, from initial offline training (T=0) to three months post-deployment reflecting environmental evolution, including occupancy changes and signal interference.

TABLE I. DETAILS OF SMARTPHONES USED IN EXPERIMENTS

| Manufacturer | Model | Acronym | Wi-Fi Chipset |
|---|---|---|---|
| BLU | Vivo 8 | BLU | MediaTek Helio P10 |
| HTC | U11 | HTC | Qualcomm Snapdragon 835 |
| Samsung | Galaxy S7 | S7 | Qualcomm Snapdragon 835 |
| LG | V20 | LG | Qualcomm Snapdragon 820 |
| Motorola | Z2 | MOTO | Qualcomm Snapdragon 820 |
| OnePlus | 3 | OP3 | Qualcomm Snapdragon 820 |

The performance of *DAILOC* was assessed using the Euclidean distance (ED) as defined in (6):

$$ED = \sqrt{(X_{pred} - X_{true})^2 + (Y_{pred} - Y_{true})^2 + (Z_{pred} - Z_{true})^2} \quad (6)$$

ED was used to measure the distance between predicted location $(X_{pred}, Y_{pred}, Z_{pred})$ and ground-truth location $(X_{true}, Y_{true}, Z_{true})$, offering a direct and interpretable metric for evaluating localization accuracy, where smaller distance values indicate better performance.

### B. Comparison with state-of-art

Fig. 5 compares the mean localization error of *DAILOC* against three state-of-the-art indoor localization frameworks: DARE [27], FIDORA [21], and FEDHIL [18] (all discussed earlier in Section II). In this experiment, one new device is introduced sequentially across different time instances for supervised onboarding. Once onboarded, each device undergoes unsupervised adaptation, with the study spanning a duration of three months. This setup enables a comprehensive evaluation of *DAILOC*'s capabilities across three key aspects of plasticity, stability and temporal robustness.

The results in Fig. 5 are averaged across Building 1 and Building 2 across the six time instances, from deployment time (T=0) to three months post-deployment. While all frameworks perform comparably at T=0, their performance diverges over time as domain shifts emerge. DARE [27], a domain-incremental learning-based method, demonstrates strong plasticity and effectively integrates new domains. However, in the context of indoor localization where ground truth is unavailable post-deployment, it relies on pseudo-labels for unsupervised learning. This leads to increased susceptibility to pseudo-label noise, reducing its stability over time. FIDORA [21], by contrast, is a domain adaptation approach focused on aligning features without retraining. It maintains better stability over time but lacks plasticity, as it cannot incorporate new devices or domain variations incrementally. FEDHIL [18], a federated learning-based method, provides moderate early performance but is prone to catastrophic forgetting in the long term. As a result, its accuracy degrades steadily over time. In contrast, *DAILOC* combines domain-incremental learning with domain-aware disentanglement and memory-guided class latent alignment. It preserves plasticity for integrating new devices and maintains stability during unsupervised adaptation, resulting in consistently lower error and reduced variance over time.

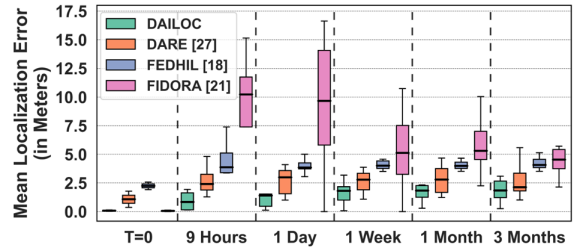

Fig. 5 Comparison of *DAILOC* against state-of-the-art.

### C. DAILOC Performance Analysis

In this next analysis, we present a more detailed set of results for our framework. Fig. 6 and 7 present mean localization error (in meters) across devices and time instances for Building 1 and Building 2 respectively, for *DAILOC*. Each heatmap reflects how localization error evolves as devices are incrementally introduced (as in the previous experiment) and the system adapts over time in an unsupervised manner.

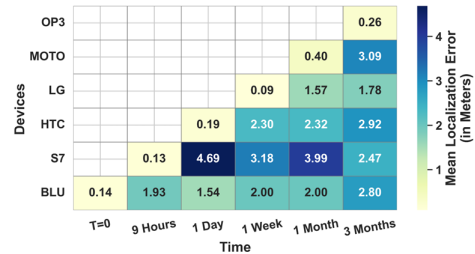

Fig. 6 Comparison of mean localization error (in meters) for Building 1

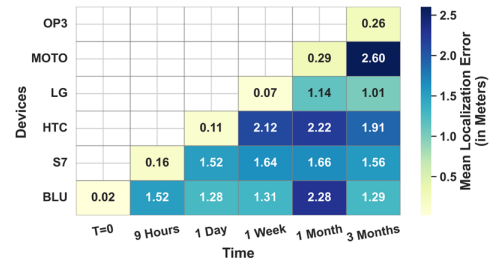

Fig. 7 Comparison of mean localization error (in meters) for Building 2

*DAILOC* demonstrates strong localization performance in both buildings, despite their architectural and signal distribution differences. For instance, error remains under 2m for most devices even after 3 months of domain drift. In Building 2, which features more complex signal reflections due to metal shelving and open structures, *DAILOC* maintains performance more effectively than in Building 1. For example, LG maintains a low error from week 1 to month 3 (1.14 m – 1.01 m), indicating resilience to longer-term drift.

While devices like S7 and BLU show higher errors in Building 1 after longer time periods (e.g., S7: 3.99m error at month 1), *DAILOC*'s overall performance remains significantly more stable compared to state-of-the-art methods (see Fig. 5). These results across both buildings highlight that *DAILOC*'s architectural design, rooted in disentangled representation and domain-incremental learning, is effective not only in adapting to new devices, but also in sustaining performance across varying spatial contexts and over time.

*D. Ablation Study: Impact of Disentanglement*

To evaluate the role of disentangled representation learning on pseudo-label noise, we measure the mean pseudo-label error before and after disentanglement across five time instances in both buildings. As shown in Fig. 8, applying disentanglement significantly lowers pseudo-label error, especially in early time steps where domain shifts are more severe and representations are poorly adapted. In Building 1, the error drops by over 80% at 9 hours (from 11.0 m to 1.5 m), and remains consistently lower throughout. A similar trend is observed in Building 2, with pseudo-label noise reduced by up to 60% across time instances. These results confirm that disentangling domain-specific features from location-relevant ones improves the reliability during unsupervised adaptation, making the framework more robust to noisy pseudo labels.

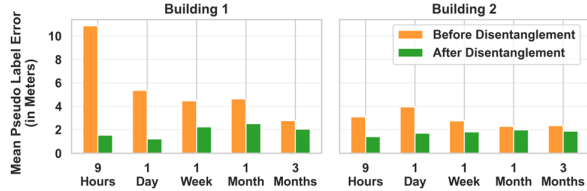

Fig. 8 Mean pseudo-label error before and after disentanglement across time in two buildings.

*E. Ablation Study: Impact of Device Onboarding Order*

To evaluate robustness to the order of device introduction over time, we randomly shuffled the sequence of introducing new devices, to eliminate any bias in our results. As shown in Fig. 9, across 5 different device orderings over time, *DAILOC* consistently achieves low localization error, demonstrating resilience to variability in heterogeneous device arrival. In contrast, state-of-the-art methods such as DARE and FEDHIL show moderate sensitivity, while FIDORA exhibits the highest degradation due to its inability to preserve stability across sequential domains, leading to catastrophic forgetting when devices arrive unpredictably.

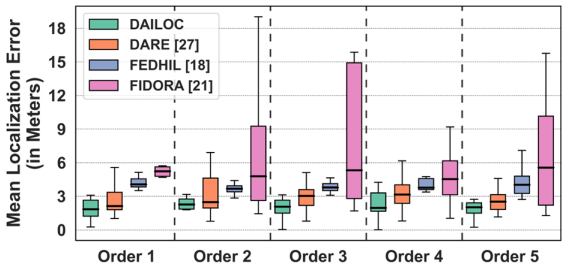

Fig. 9 Mean localization error (in meters) across different heterogeneous device introduction orderings over time

## VI. CONCLUSION

In this paper, we introduced *DAILOC*, a domain-incremental learning framework for robust Wi-Fi fingerprinting-based indoor localization under real-world conditions. By disentangling domain-specific and location-relevant features using a ML-VAE, and leveraging domain-aware reparameterization with memory-guided class latent alignment, *DAILOC* effectively adapts to both temporal/environmental and device-induced shifts. Evaluations across multiple buildings, smartphones, and deployment stages show that *DAILOC* consistently outperforms the state-of-the-art, achieving 2.74× lower average error and up to 4.6× lower worst-case error.